\documentclass[letterpaper, 10 pt, journal, twoside]{ieeetran}  % Comment this line out if you need a4paper
\IEEEoverridecommandlockouts                              % This command is only needed if 
\usepackage{graphics} 
\usepackage{epsfig} 
\usepackage{mathptmx} 
\usepackage{times} 
\usepackage{amsmath} 
\usepackage{amssymb} 
\graphicspath{ {fig/} }
\usepackage{algorithmic}
\usepackage{algorithm}
\usepackage{url}
\usepackage{subcaption}
\usepackage{graphicx}
\usepackage[dvipsnames]{xcolor}
\usepackage{balance} % balance reference list on last page
\usepackage{hyperref}
\usepackage{float}
\usepackage[noadjust,compress]{cite}

\title{First do not fall: learning to exploit  a wall\\ with a damaged humanoid robot
}

\author{Timothée Anne$^{1}$, Eloïse Dalin$^{1}$, Ivan Bergonzani$^{1}$, Serena Ivaldi$^{1}$, and Jean-Baptiste Mouret$^{1}$%
\thanks{Preprint June, 2022. Full reference: Anne et al. (2022). "First do not fall: learning to exploit a wall with a damaged humanoid robot". In Robotics and Automation Letters.}%Use only for final RAL version
\thanks{Experiments presented in this paper were carried out using the Grid'5000 testbed, supported by a scientific interest group hosted by Inria and including CNRS, RENATER and several Universities as well as other organizations (see \url{https://www.grid5000.fr}). This project is supported by the CPER SCIARAT, the CPER CyberEntreprise, the Direction General de l'Armement (convention Inria-DGA ``humanoïde résilient''), and the Creativ'Lab platform of Inria/LORIA.} %Use only for final RAL version
\thanks{$^{1}$Université de Lorraine, CNRS, Inria Nancy - Grand Est. Contact: jean-baptiste.mouret@inria.fr}
}
\begin{document}
\bstctlcite{IEEEexample:BSTcontrol}

\maketitle
%\thispagestyle{empty}
%\pagestyle{empty}

%%%%%%%%%%%%%%%%%%%%%%%%%%%%%%%%%%%%%%%%%%%%%%%%%%%%%%%%%%%%%%%%%%%%%%%%%%%%%%%%
%   Abstract
%%%%%%%%%%%%%%%%%%%%%%%%%%%%%%%%%%%%%%%%%%%%%%%%%%%%%%%%%%%%%%%%%%%%%%%%%%%%%%%%
\begin{abstract}
Humanoid robots could replace humans in hazardous situations but most of such situations are equally dangerous for them, which means that they have a high chance of being damaged and falling. We hypothesize that humanoid robots would be mostly used in buildings, which makes them likely to be close to a wall. To avoid a fall, they can therefore lean on the closest wall, as a human would do, provided that they find in a few milliseconds where to put the hand(s). This article introduces a method, called D-Reflex, that learns a neural network that chooses this contact position given the wall orientation, the wall distance, and the posture of the robot. This contact position is then used by a whole-body controller to reach a stable posture. We show that D-Reflex allows a simulated TALOS robot (1.75m, 100kg, 30 degrees of freedom) to avoid more than 75\% of the avoidable falls and can work on the real robot.
\end{abstract}

%%%%%%%%%%%%%%%%%%%%%%%%%%%%%%%%%%%%%%%%%%%%%%%%%%%%%%%%%%%%%%%%%%%%%%%%%%%%%%%%
%   Introduction
%%%%%%%%%%%%%%%%%%%%%%%%%%%%%%%%%%%%%%%%%%%%%%%%%%%%%%%%%%%%%%%%%%%%%%%%%%%%%%%%
\section{Introduction}

\IEEEPARstart{H}{umanoid} robots are some of the most versatile machines ever designed \cite{humanoids_ref}. They can grasp, pull, push, hold, and reach for both low or high places, but they can also walk, climb stairs, or crawl in a tunnel. Thanks to their small footprint, they can navigate in narrow spaces, and, more generally in all the environments designed for humans.

This versatility makes humanoids ideal machines to be deployed in risky and complex situations, like industrial disasters or space operations, during which more versatility means a higher probability of having the right set of capabilities to solve the problem at hand \cite{humanoids_ref,DRC_paper}. This contrasts with industrial robots, which are designed to perform the same set of tasks continuously in a well-defined environment.

Nevertheless, the versatility of humanoid robots comes at the cost of an increased fragility: they have more joints than most robots and a single joint failure often results in a fall \cite{DRC_paper}. This fragility is especially concerning because humanoid robots would be the most useful in situations that are too dangerous for humans, which are likely to be equally dangerous for a robot. A deployed humanoid robot is therefore likely to be damaged during some of its missions.

\begin{figure}[t]
    \centering
    \includegraphics[width=1.\linewidth]{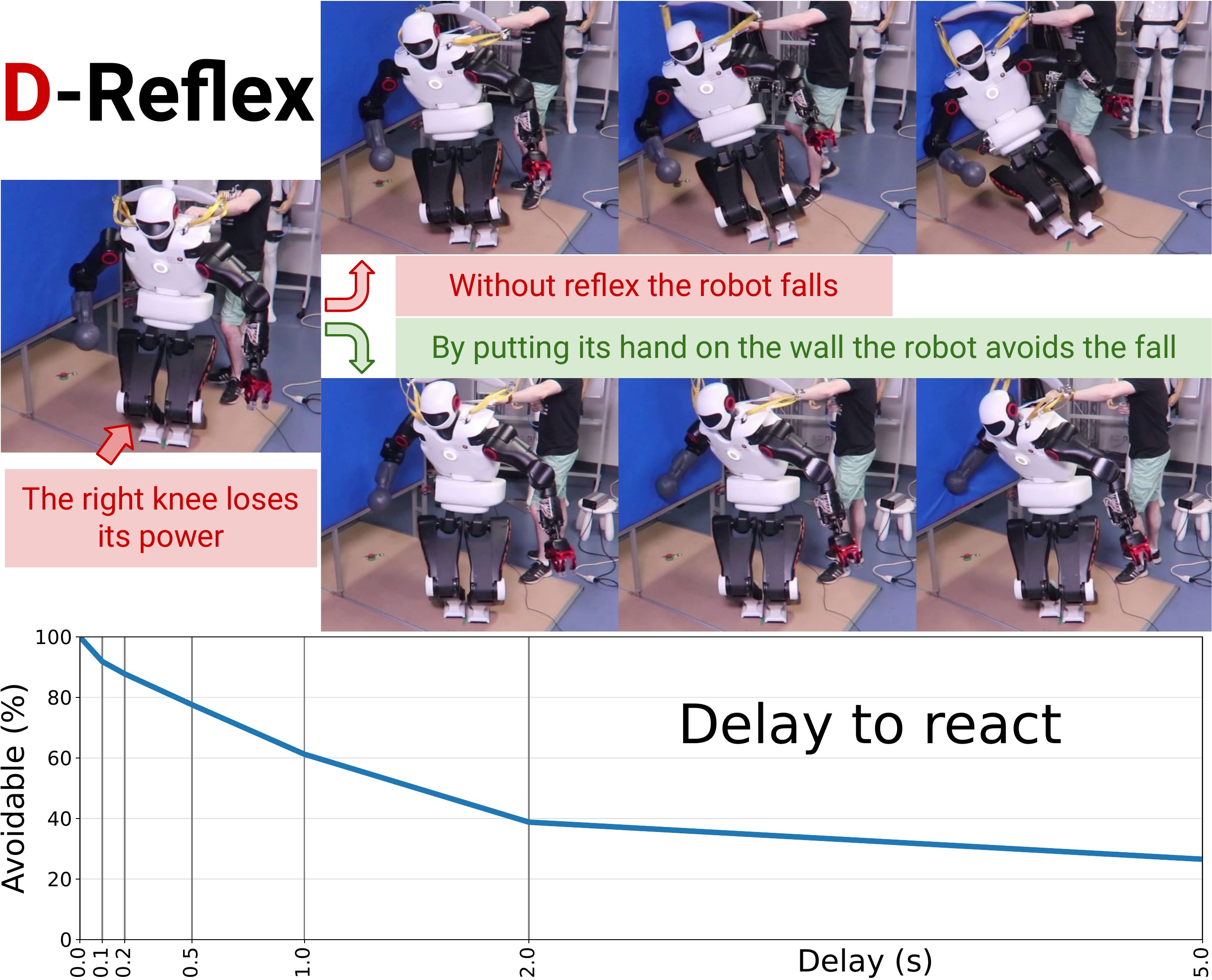}
    \caption{(top) A humanoid robot detects a fault on one of its legs. If it does nothing, it falls; but it can recover its stability by putting its ``hand'' on the wall at the appropriate location (depending on its posture,  the wall distance, and the wall orientation). (bottom) The percentage of avoidable falls (see Sec.~\ref{sec:avoidable} for the definition of ``avoidable'') decreases when the delay taken between damage and reflex increases; the robot has only a few milliseconds to react with a high success rate.}
    \label{fig:context}
    \vspace{-0.5cm}
\end{figure}

The traditional approach for robot damage recovery is to identify the damage, update the model, then use the updated model to perform the tasks \cite{model_identification}. Unfortunately, a falling humanoid robot has only a few milliseconds to make a decision (Fig.~\ref{fig:context}), whereas identifying the dynamical model of such a highly redundant robot requires extensive data and specific trajectories \cite{model_identification,talos_actuator_identification}. A few learning-based damage recovery algorithms for 2-legged \cite{qp_repulsors}, 4-legged \cite{mine_meta_learning, rituraj_meta_learning} or 6-legged robots \cite{nature_paper, nagabandi_meta_learning, residual_paper2} were recently published, but they all assume that the robot can try a behavior, fall, and try again until it finds a compensatory behavior. Humanoids usually cannot afford to fall, and trial-and-error would require an ability to stand up without a perfect knowledge of the robot model, which is very challenging.

Given the envisioned missions, humanoid robots are more likely to be used in indoor environments, for instance in damaged buildings, than in open fields. As a result, these robots are likely to have walls or furniture within reach: taking inspiration from human reflexes, a promising strategy in case of leg damage is, therefore, to lean on the nearest wall to avoid the fall. 

In this article, our main contribution is a method that allows damaged humanoid robots to avoid many falls by leaning on a nearby wall (Fig. \ref{fig:context}). Once the robot is stable, another algorithm could update the model and activate compensatory behaviors (e.g., walking on one leg using the wall). Our main assumptions are that (1) we can detect the occurrence of a fault in the leg, but we do not know its origin (missing parts, disconnected from power, or locked); meaning that we do not know the exact model of the damaged robot and (2) we know the distance and the angle to the nearest wall (for instance with a dedicated sensor on the shoulder). With these assumptions, our method, called D-Reflex (for ``Damage-Reflex''), finds in a few milliseconds where to put the hand on the wall, and reaches a stable state using a whole-body controller \cite{bouyarmane2017weight,dalin_whole-body_2021}.

The main idea of D-reflex is to learn a neural network that predicts the success chance of each potential contact location on the wall given a posture and a wall configuration. This neural network is learned with supervised learning using a dataset created by simulating many situations and many potential contact positions. Splitting the process into these two steps---creating a dataset in simulation, then learning a predictor---makes it possible to exploit the mature and well-understood supervised learning algorithms while separating the learning process from the simulation process. Once learned, the neural network can be queried in a few milliseconds, which is ideal for a fast emergency reflex while the robot is falling. 

We report experiments with both a simulated and a real TALOS humanoid robot (1.75m, 100kg, 30 degrees of freedom)~\cite{talos_ref} with a leg damaged in several ways.

%%%%%%%%%%%%%%%%%%%%%%%%%%%%%%%%%%%%%%%%%%%%%%%%%%%%%%%%%%%%%%%%%%%%%%%%%%%%%%%%
%   Related Work
%%%%%%%%%%%%%%%%%%%%%%%%%%%%%%%%%%%%%%%%%%%%%%%%%%%%%%%%%%%%%%%%%%%%%%%%%%%%%%%%
\section{Related Work}

%%%%%%%%%%%%%%%%%%%%%%%%%%%%%%%%%%%%%%%%%%%%%%%%%%%%%%%%%%%%
%%   Multi-contact planning and damage identification
%%%%%%%%%%%%%%%%%%%%%%%%%%%%%%%%%%%%%%%%%%%%%%%%%%%%%%%%%%%%
\subsection{Multi-contact planning and damage identification}
Multi-contact planning, for instance choosing to put a hand on a wall, has been a long-lasting research topic in the humanoid robotics community (e.g., \cite{kheddar2019humanoid, PolveriniHLT21}). However, planning algorithms tend to require more than the few milliseconds that a falling humanoid robot can afford (e.g. about 100ms for a known target point \cite{Crocoddyl}). In addition, all planning algorithms assume a known model of the dynamics, which is not the case when the robot is damaged, whereas the dynamics of the fall are typically very different with a damaged robot. 

A preliminary step to any planning algorithm is therefore to identify the model of the damaged robot \cite{model_identification}. Unfortunately, identifying the dynamical model of a humanoid requires at least several minutes of data collection and specific ``exciting'' trajectories. For example, to identify the TALOS robot's \cite{talos_ref} elbow joint, Ramuzat \textit{et al.} \cite{talos_actuator_identification} required $230s$ of data collection and a well-chosen exciting trajectory. It seems highly optimistic to be able to identify a new model with a few milliseconds of sensor data. Overall, we see the present contribution as a preliminary step to a system identification step: the robot takes an emergency decision, then, once stable, it can get data to identify the new model and rely on model-based planning or control algorithms with an updated model.

%%%%%%%%%%%%%%%%%%%%%%%%%%%%%%%%%%%%%%%%%%%%%%%%%%%%%%%%%%%%
%%   Fall and damage mitigation
%%%%%%%%%%%%%%%%%%%%%%%%%%%%%%%%%%%%%%%%%%%%%%%%%%%%%%%%%%%%
\subsection{Fall and damage mitigation}

Inspired by how humans react when falling in front of a wall, Cui \textit{et al.} \cite{cui_human_2021} recently proposed to move the robot arms in a way that maximizes the ellipsoid stiffness, thus increasing stability and shock absorption. Compared to the present work, Cui \textit{et al.} only worked on collisions with a frontal wall and, more importantly, with an intact robot whose model is known.

When the fall is inevitable, several methods have been demonstrated on different humanoids to mitigate the damage.
During the ``pre-impact'' phase, the robot adapts its posture by avoiding hand-designed ``fall singularities'' postures that increase the impact \cite{kheddar1}, by seeking to take a safe posture when falling on the back \cite{fall_minimization_back}, or, by performing a rollover strategy \cite{Liu21}.
During the ``impact-time'' phase, the  PD-gains are automatically adapted by incorporating them in the QP-controller to prevent the actuators from reaching their torque limits while still reaching the desired posture \cite{kheddar2}. 
During the ``post-impact'', a force distribution quadratic program distributes the exceeding linear momentum gathered during fall into the different body parts \cite{kheddar3}.  
These methods do not try to avoid the fall but mitigate the damage induced by the fall. Overall, this line of work is complementary to ours: while many falls can be avoided by adding a contact to the wall, some falls cannot be avoided in that way and the only behavior left is to mitigate the damage.

%%%%%%%%%%%%%%%%%%%%%%%%%%%%%%%%%%%%%%%%%%%%%%%%%%%%%%%%%%%%
%%   Machine learning
%%%%%%%%%%%%%%%%%%%%%%%%%%%%%%%%%%%%%%%%%%%%%%%%%%%%%%%%%%%%
\subsection{Machine learning for damage recovery and mitigation}
A promising approach to allow robots to adapt to unforeseen damage is to learn dynamical models with generic machine learning methods, like neural networks or Gaussian processes. To minimize the amount of data required, current methods leverage a simulation of the intact robot \cite{residual_paper2} or meta-learning to train the model for fast adaptation \cite{nagabandi_meta_learning, rituraj_meta_learning, mine_meta_learning}. Successful experiments with 4-legged \cite{rituraj_meta_learning, mine_meta_learning} and 6-legged \cite{residual_paper2, nagabandi_meta_learning} have been reported, but, to our knowledge, not with 2-legged robots. One important difference between humanoids and other multi-legged robots is that the latter can afford to fall during learning because getting up is easy.

Spitz \textit{et al.} \cite{qp_repulsors} investigated how a damaged humanoid robot can adapt its behavior by avoiding states that previously led to a fall. Similarly, Cully et al. \cite{nature_paper} designed a learning algorithm that allows a damaged 6-legged robot to find a new behavior that does not rely on the damaged parts. In both cases, the authors used an episodic setup in which the robot can try a behavior, fall, and try again many times. This setup is not applicable when the objective is to avoid the fall immediately after damage.

To our knowledge, the closest work to ours is for a quadruped robot that performs ``cat-like'' acrobatics to land on its feet when falling from several meters~\cite{mini_cheetah_falling}. To solve this problem, the authors used a trajectory optimization algorithm to generate a dataset of examples in simulation, then trained a neural network policy that imitates these reflexes but is fast enough to be used on the robot. Similarly, the D-reflex approach first solves the problem with extensive simulations, then uses these results to train a fast neural network that selects the most appropriate behavior.

%%%%%%%%%%%%%%%%%%%%%%%%%%%%%%%%%%%%%%%%%%%%%%%%%%%%%%%%%%%%%%%%%%%%%%%%%%%%%%%%
%   Clear formulation of the problem
%%%%%%%%%%%%%%%%%%%%%%%%%%%%%%%%%%%%%%%%%%%%%%%%%%%%%%%%%%%%%%%%%%%%%%%%%%%%%%%%
\section{Problem}

%%%%%%%%%%%%%%%%%%%%%%%%%%%%%%%%%%%%%%%%%%%%%%%%%%%%%%%%%%%%
%%   Formulation
%%%%%%%%%%%%%%%%%%%%%%%%%%%%%%%%%%%%%%%%%%%%%%%%%%%%%%%%%%%%

\subsection{Considered Damage Conditions}

A robot can be damaged in many ways and we often cannot determine them precisely. We use three damage conditions: amputation (missing parts), passive actuators (the actuators do not deliver any torque, i.e., the joints can turn freely in the corresponding axis), and locked actuators (the joint position is fixed). The first two damage conditions are critical for the stability of the robot and very often result in a fall when applied to one of the legs, but they cannot be distinguished by querying the control board of each joint: in both cases, the control boards do not answer or send an error.

To show that our method is robust to different combinations of damage conditions, we sample different combinations of those three damage conditions on the six joints of the leg. We call $J$ the set of damaged joints.

\subsection{Formulation}

A humanoid robot suffers an unexpected damage combination on one of its legs which would often result in a fall if nothing is done. The goal is to find a contact position on the wall that allows the robot to avoid the fall and stay stable (Fig.~\ref{fig:context}).

We know:
\begin{itemize}
    \item that one of the legs is damaged and which one: we assume that the faulty joints control boards do not answer or return an error;
    \item the presence of a plane wall within arm's reach, its distance $d$ and its orientation $\alpha$ with regard to the robot (this position can be easily known with a RGB-D sensor on the shoulder of the robot and fitting a plane to the point cloud \cite{RANSAC});
    \item the current posture $q$ of the robot when the fault is detected.
\end{itemize}

We do not know:
\begin{itemize}
    \item the set of damaged joints $J$ and the nature of the damage condition: the leg could be fully amputated, but it could also have lost the power;
    \item the successful contact positions on the wall (there are usually several possible successful contact positions).
\end{itemize}

We want to learn a function $f: (q, d, \alpha) \rightarrow (x^*,y^*)$ that returns a successful contact position \emph{robust} to the nature of the damage condition. We assume that the robot has a controller that can make it move to reach this position.

%%%%%%%%%%%%%%%%%%%%%%%%%%%%%%%%%%%%%%%%%%%%%%%%%%%%%%%%%%%%%%%%%%%%%%%%%%%%%%%%
%   Method
%%%%%%%%%%%%%%%%%%%%%%%%%%%%%%%%%%%%%%%%%%%%%%%%%%%%%%%%%%%%%%%%%%%%%%%%%%%%%%%%
\section{Method}

\begin{figure*}[t]
\vspace*{5mm}
    \centering
    \includegraphics[width=0.98\linewidth]{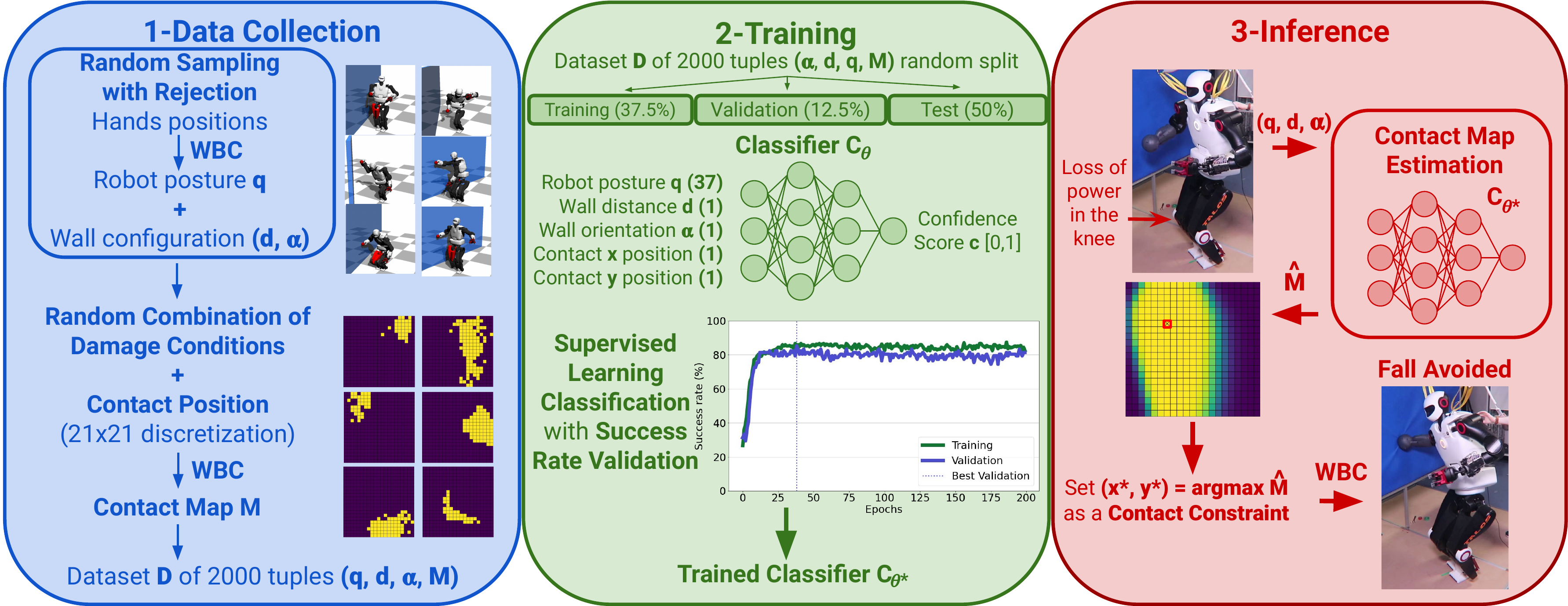}
    \caption{Overview of D-Reflex. In the first step (Data Collection), we sample different situations: wall configurations (distance and orientation), robot postures, and damage combinations, and simulate the behavior of the robot for each possible point on a 21$\times$ 21 grid on the wall (441 simulations for each situation). In the second step, we train a neural network classifier to predict the success (avoiding the fall) for each point of the grid. In the third step, we query the trained neural network for each point of the grid to select the best contact position, and we set it as a contact constraint in the whole-body controller.}
    \label{fig:method_schematic}
    \vspace{-0.5cm}
\end{figure*} 

%%%%%%%%%%%%%%%%%%%%%%%%%%%%%%%%%%%%%%%%%%%%%%%%%%%%%%%%%%%%
%%   Whole-body Control
%%%%%%%%%%%%%%%%%%%%%%%%%%%%%%%%%%%%%%%%%%%%%%%%%%%%%%%%%%%%
\subsection{Whole-body Control}
The humanoid robot is controlled with a whole-body controller (WBC) based on quadratic programming \cite{bouyarmane2017weight,dalin_whole-body_2021}. At each time step, the robot searches for the acceleration of each joint that minimizes a sum of quadratic cost functions constrained by the dynamical model of the robot (an equality constraint) and other equality/inequality constraints:

\begin{equation}
    \begin{array}{l}
         (\tau^*, \ddot{q}^*) = \underset{\tau,\ddot{q}}{argmin} \underset{k=0}{\overset{n\_tasks}{\sum}} w_k \|A_k(q,\dot{q})\ddot{q}-b_k(q,\dot{q})\|^2 \\ 
         \text{ s.t. }  A_{ineq}(q,\dot{q})\ddot{q}\leq b_{ineq}(q,\dot{q})\\
         \text{ s.t. }  A_{eq}(q,\dot{q})\ddot{q}=b_{eq}(q,\dot{q})\\
         \text{ s.t. }  \tau = M(q)\Ddot{q} + F(q,\Dot{q})\\
    \end{array}
\end{equation}
where $\tau^*$ and $\ddot{q}^*$ are the optimal joint torques and accelerations, $w_k$ is the weight of the task $k$ described by $A_k(q,\dot{q})$ and $b_k(q,\dot{q})$, $M$ is the joint-space mass matrix and F contains all the non-linear terms (Coriolis, centrifugal, gravity, and contacts).
    
In this work, we used the controller described in Dalin et al. \cite{dalin_whole-body_2021}. The main tasks are the Cartesian position and orientation of the hands and feet, the position of the center of mass, and a default postural task which is used to ensure the unicity of the QP problem and bias towards a ``neutral'' position. The constraints are the feet contact with the floor and the joint position, velocity, and acceleration bounds. 

Once the optimal torque and joint accelerations are computed, we integrate them using the model of the robot to get the desired joint positions, which we pass to the low-level joint controllers. In the real TALOS robot, the joints controllers are PID controllers implemented on Ethercat boards~\cite{talos_ref}. In simulation, we use a Stable-PD controller \cite{SPD} that leverages the model of the robot to compute torques given a position target, which is in our experience a good approximation of well-tuned PID controllers.

Please note that the model used by the controller is not updated after the damage because we do not know the extent of the damage condition. As a result, the solution needs to be robust to different combinations of damage conditions.

%%%%%%%%%%%%%%%%%%%%%%%%%%%%%%%%%%%%%%%%%%%%%%%%%%%%%%%%%%%%
%%   Data Collection
%%%%%%%%%%%%%%%%%%%%%%%%%%%%%%%%%%%%%%%%%%%%%%%%%%%%%%%%%%%%
\subsection{Data Collection}

The first part of the process consists in generating random configurations of the wall and postures of the robot, and collecting the corresponding truth value of contact positions (Fig.~\ref{fig:method_schematic}, part 1).  

%%%%%%%%%%%%%%%%%%%%%%%%%%%%%%%%%%%%%%%%%%%%%%
%%%   Sampling random configuration
%%%%%%%%%%%%%%%%%%%%%%%%%%%%%%%%%%%%%%%%%%%%%%
\subsubsection{Sampling random configuration}
To randomly sample plausible random postures of the robot, we first randomly sample target hands positions in a cuboid facing the robot. We set these targets in our whole-body controller as Cartesian tasks and let it run for 4s which makes the robot take a posture $q$. Each random hands positions induces a different but feasible posture, such as knee bending to reach a lower position or rotated torso with arms raised.  

%%%%%%%%%%%%%%%%%%%%%%%%%%%%%%%%%%%%%%%%%%%%%%
%%%   Sampling rejection
%%%%%%%%%%%%%%%%%%%%%%%%%%%%%%%%%%%%%%%%%%%%%%
\subsubsection{Sampling wall positions}
We need random wall configurations (distance $d$ and orientation $\alpha$) that are within arm's reach but that do not collide during the initial motion to reach the target posture (because this would make the robot fall before any damage). To ensure this property, we exclude wall configurations that collide during the first $4s$ of motion.

%%%%%%%%%%%%%%%%%%%%%%%%%%%%%%%%%%%%%%%%%%%%%%
%%%   Grid search
%%%%%%%%%%%%%%%%%%%%%%%%%%%%%%%%%%%%%%%%%%%%%%
\subsubsection{Construction of contact maps}
Once we have the posture of the robot $q$ and the wall configuration (distance $d$ and orientation $\alpha$), we discretize the wall's plane and run the simulation for each considered contact position by (1) reaching the posture $q$, (2) damage the robot in the simulation after $4s$ without waiting for the robot to be stable, i.e. the current momentum and angular momentum induced by the motion is kept, (3) add a Cartesian contact constraint in our whole-body controller to ask the hand to be in contact at the given position, and (4) let the simulation run for $11s$. A contact position is deemed correct if at the end of this 15s-episode the robot has not touched the floor with anything else than its feet, or the wall with anything else than its hand. We only consider contacts with the hand because it allows clean contacts with known interaction forces (the wrist has a force-torque sensor) and can reach the farthest positions to make contact as soon as possible. This gives us a Boolean matrix $M\in\{0,1\}^{(21,21)}$ that we call ``Contact Map''. Fig.~\ref{fig:method_schematic}.1 shows some examples.   

%%%%%%%%%%%%%%%%%%%%%%%%%%%%%%%%%%%%%%%%%%%%%%%%%%%%%%%%%%%%
%%   Training
%%%%%%%%%%%%%%%%%%%%%%%%%%%%%%%%%%%%%%%%%%%%%%%%%%%%%%%%%%%%
\subsection{Training}
\label{sec:training}
Once we have a dataset $D$ of samples $(q,d,\alpha, M)$ split into a training set (37.5\%), a validation set (12.5\%), and a test set (50\%), we want to learn the function 
$$
f: (q, d, \alpha) \rightarrow (x^*,y^*)
$$
where $(x^*,y^*)$ is a successful contact position in the wall referential. As the contact map $M$ often contains several successful contact positions, we cannot directly learn $f$. 

Instead, we learn the classification function that predicts the success of each input point:
$$ 
C: (q, d, \alpha, x, y) \rightarrow c
$$
where $c\in[0,1]$ represents the confidence of the contact position $(x,y)$ to be a successful contact position. 

To estimate this classification function $C$, we train a neural network
$\hat{C}_{\theta}$ parameterized by the set of weights $\theta$ using PyTorch with Cross-Entropy error and the Adam optimizer. After performing an extensive random search on the hyper-parameters, we have selected a fully-connected feed-forward neural network with 2 hidden layers of 1024 units with ReLU activation function, a dropout of 0.2, and a learning rate of $10^{-5}$. We obtained similar results with other combinations of network architecture, cost function, and optimizer.

To get an estimation of the function $f$, we query the neural network with points $(x,y) \in (X,Y)$ on the wall and select the point with the highest activation:

$$
\hat{f}(q,d,\alpha) = \underset{(x,y) \in (X,Y)}{argmax} \hat{C}_{\theta}(q,d,\alpha,x,y).
$$

Since the points are 2-dimensional and there is no need for a millimeter-scale precision, this optimization can be effectively performed using a straightforward grid search. In that case, it typically requires a few hundred calls to the neural network (e.g., $441$ calls for a $21\times21$ grid), which is easily done in parallel on modern GPUs (about 4ms for 441 queries on an Nvidia RTX 2080) and CPUs (about 17ms on the Talos's CPU).

In supervised learning, we would periodically evaluate the prediction score of the neural network on the validation set to avoid overfitting and select the final set of weights. Here, the utility of our learned neural network is measured by the effectiveness of the selected contact position, which we evaluate by running a simulation with a damaged robot. In other words, we evaluate the neural network by looking at $\hat{f}(q,d,\alpha)$ on the validation set instead of the traditional $\hat{C}_{\theta}(q,d,\alpha,x,y)$ because we want to evaluate the ``end product'' and not the neural network itself.  At the end of the optimization, we select the set of weights $\theta$ that corresponds to the highest success rate on the validation set (Figure~\ref{fig:method_schematic}.2).

%%%%%%%%%%%%%%%%%%%%%%%%%%%%%%%%%%%%%%%%%%%%%%%%%%%%%%%%%%%%
%%   Inference
%%%%%%%%%%%%%%%%%%%%%%%%%%%%%%%%%%%%%%%%%%%%%%%%%%%%%%%%%%%%
\subsection{Inference}

During its mission, the robot performs its tasks and detects damage in one of its legs by querying the boards periodically. We get the last known posture of the robot $q$, the distance $d$, and orientation $\alpha$ to the closest wall. For this preliminary work, we hypothesize that the robot can evaluate $\alpha$ and $d$ with a dedicated sensor.

We query the learned model $\hat{C}_\theta$ for a grid of points to estimate the best contact position $(x^*, y^*) = \hat{f}(q,d,\alpha)$. Fig~\ref{fig:main_figure} shows examples of predicted contact maps. We then add a contact constraint in the whole-body controller so that it puts its hand on the wall to recover its balance (Fig.~\ref{fig:method_schematic}.3). 

%%%%%%%%%%%%%%%%%%%%%%%%%%%%%%%%%%%%%%%%%%%%%%%%%%%%%%%%%%%%
%%   Implementation
%%%%%%%%%%%%%%%%%%%%%%%%%%%%%%%%%%%%%%%%%%%%%%%%%%%%%%%%%%%%
\subsection{Implementation}

%%%%%%%%%%%%%%%%%%%%%%%%%%%%%%%%%%%%%%%%%%%%%%
%%%   Triggering the reflex
%%%%%%%%%%%%%%%%%%%%%%%%%%%%%%%%%%%%%%%%%%%%%%
\subsubsection{Triggering the reflex}
We assume that the robot knows when an actuator is faulty so that we can instantly trigger the reflex. We put the robot in an ``urgent mode'' that removes the non-essential tasks: we only keep the feet contact constraints, the joints bounds constraints, the position task of the Center of Mass (CoM), and the postural task. Finally, we add the hand target contact at the selected location on the wall and let the whole-body controller generate the trajectory to reach it.

The CoM task requires a Cartesian position. By default, the target position is between the two feet. In our case, we hypothesize that a single leg is working, but, since we want a contact on the wall, the new CoM target should be somewhere between the remaining foot and the contact point. However, we do not know the real contact position because the model is incorrect. After some preliminary experiments, we have decided to not change the CoM target as it has shown good results: if we put the CoM target on the remaining foot or too close to the wall the success rate declined significantly. A better solution would be to learn the CoM target in addition to the hand contact position, which we will explore in future work. 

%%%%%%%%%%%%%%%%%%%%%%%%%%%%%%%%%%%%%%%%%%%%%%
%%%   Reflex Adaption when contact occurs
%%%%%%%%%%%%%%%%%%%%%%%%%%%%%%%%%%%%%%%%%%%%%%
\subsubsection{Post-contact updates}
We empirically found that, due to the mismatch between the incorrect model and the simulated damaged robot, the desired contact point is rarely the real contact point attained by the hand. To take this into account, we use the force sensor of the wrist to detect the contact between the hand and the wall. Then, we update the contact constraint with the current robot state to stop the robot from moving its hand, which would result in the robot pushing on the wall, bouncing back, making it more likely to fall.

%%%%%%%%%%%%%%%%%%%%%%%%%%%%%%%%%%%%%%%%%%%%%%%%%%%%%%%%%%%%%%%%%%%%%%%%%%%%%%%%
%   Experiments
%%%%%%%%%%%%%%%%%%%%%%%%%%%%%%%%%%%%%%%%%%%%%%%%%%%%%%%%%%%%%%%%%%%%%%%%%%%%%%%%
\section{Experiments}

\begin{figure*}[t]
    \centering
    \includegraphics[width=0.98\linewidth]{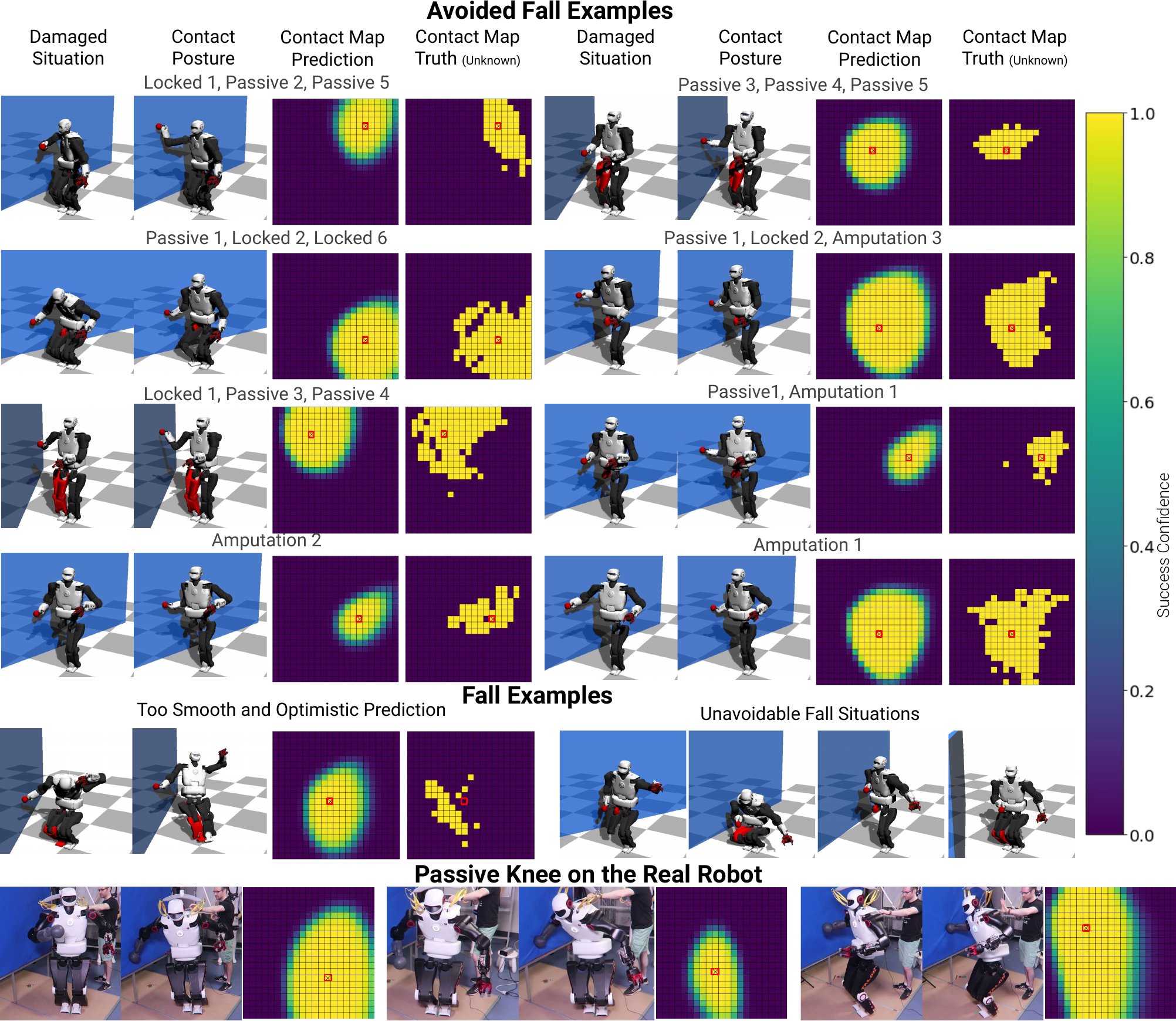}
    \caption{Examples of damaged situations and corresponding contact or falling postures using D-Reflex on the TALOS~\cite{talos_ref} robot in simulation and on the real robot, the video \url{https://youtu.be/hbuWr-ZNAtg} shows different examples. We also show the contact map estimated by our neural network and the true contact map measured but unknown during training. We distinguished two kinds of failures: when the predicted contact map is too smooth and optimistic compared to the truth, and when there is no successful contact point, i.e., the fall is unavoidable by seeking a contact on the wall.}
    \label{fig:main_figure}
    \vspace{-0.5cm}
\end{figure*}

We control 30 degrees of freedom of the TALOS robot~\cite{talos_ref}: 7 for each arm, 6 for each leg, 2 for the torso, and 2 for the head. We replaced the gripper with a ball in our simulation to be close to the experiments on the real robot (to avoid breaking the gripper), but similar results are obtained with a simulated gripper. The robot is simulated using the DART physics library \cite{lee2018dart} through the RobotDART simulator\footnote{\url{https://github.com/resibots/robot_dart}}. 

The initial posture of the robot is obtained by selecting random target positions for the hands and letting the whole-body controller reach this target in 4s. The target positions are inside the cuboid: $[-0.1m, +0.2m]$ in the sagittal axis, $[-0.4m, +0.4m]$ on the frontal axis, and $[-0.5m,+0.4m]$ on the longitudinal axis. The wall distance is sampled between $0.4m$ and $1m$. The orientation is sampled in $[-1,1]$ rad. 

The contact maps of the dataset use 21 positions on the horizontal and vertical axis, which results in 441 positions. They range from $-0.75m$ and $+0.75m$ on the horizontal axis and $-0.5m$ and $+0.75m$ on the vertical axis. Fig.~\ref{fig:main_figure} shows examples of contact maps. 

We only perform the experiments on the right side of the robot (damaging the right leg and seeking contact with the right hand). The robot being perfectly symmetrical, the data and the same neural network can be used for the other side (damaging the left leg and seeking contact with the left hand). 
To check this, we trained a D-Reflex using data from the right side (Sec.~\ref{sec:training}), then used it with a damaged left leg by symmetrizing the inputs of the classifier. In $98.5\%$ of the cases, the robot is as successful as with the right side. The resulting discrepancy is likely to come from simulation artifacts.

We sample $2000$ different situations of wall configuration, robot posture, and damage combination. We let the simulation run for 4s, after which we trigger the damage combination and immediately start the reflex. We stop the simulation after 15s or until a fall is detected. For each situation, we create a discretized contact map. Overall, this dataset requires 882,000 simulations of 15s, which we run in parallel on a multicore computer in less than 3 days. 

The source code and the dataset are available online\footnote{\url{https://github.com/resibots/d-reflex}}.

%%%%%%%%%%%%%%%%%%%%%%%%%%%%%%%%%%%%%%%%%%%%%%%%%%%%%%%%%%%%
%%   Baseline
%%%%%%%%%%%%%%%%%%%%%%%%%%%%%%%%%%%%%%%%%%%%%%%%%%%%%%%%%%%%
\subsection{Baselines}

We compare against two baselines:
\begin{itemize}
    \item No Reflex: we do nothing, to highlight the importance of a reflex strategy; 
    \item Random Reflex: we randomly select the contact position on the wall, to highlight the difficulty of the problem and the necessity of learning. 
\end{itemize}

%%%%%%%%%%%%%%%%%%%%%%%%%%%%%%%%%%%%%%%%%%%%%%%%%%%%%%%%%%%%
%%   Ablations and Additions
%%%%%%%%%%%%%%%%%%%%%%%%%%%%%%%%%%%%%%%%%%%%%%%%%%%%%%%%%%%%
\subsection{Ablations and Additions}

To understand the contribution of each part we also compare against 3 ``ablation experiments'':
\begin{itemize}
    \item Posture Ablation: we remove the posture $q$ from the neural network inputs and re-train the network (this allows us to answer the question: ``does the contact position depend on the posture?'');
    \item Wall Ablation: we remove the wall configuration (distance $d$ and orientation $\alpha$) from the neural network inputs and re-train the network (``does the contact position depends on the wall configuration?'');
    \item Both Ablation: the contact point is independent of both the robot posture and the wall configuration (``is there a solution that works for any wall configuration and robot posture?''). In this case, we do not train a neural network but, instead, we use the training data to select the average best contact position.  
\end{itemize}

To check that we are not ignoring any useful information, we compare against 2 ``addition experiments'':
\begin{itemize}
    \item J Addition: we add the set of faulty joints $J$ to the neural network inputs and re-train the network (this allows us to answer the question: ``is it useful to know which joints are damaged?'');
    \item $J + \mathrm{d}q$ Addition: we add to the neural network inputs the set of faulty joints $J$ and the joints velocities $\mathrm{d}q$ and re-train the network (this allows us to answer the question: ``is it useful to know which joints are damaged and the joints velocity $\mathrm{d}q$?'').
\end{itemize}

%%%%%%%%%%%%%%%%%%%%%%%%%%%%%%%%%%%%%%%%%%%%%%%%%%%%%%%%%%%%
%%   Evaluation
%%%%%%%%%%%%%%%%%%%%%%%%%%%%%%%%%%%%%%%%%%%%%%%%%%%%%%%%%%%%
\subsection{Evaluation}
\label{sec:avoidable}
By analyzing the contact maps, we found that for about 31.5\% of the situations of the dataset there are no position of the hand on the wall that prevents the fall. Fig. \ref{fig:main_figure} shows 4 examples of such unavoidable situations. Using a decision tree classifier, we found that two criteria discriminate most of the data: the distance of the right hand to the wall and the wall orientation. If the wall is too far from the right hand the fall becomes impossible to avoid by putting the right hand on the wall.
Moreover, in many cases, only a few sparse contact positions of the $21\times21$ grid allow the robot to avoid the fall. We consider that these contact positions are unlikely to exist with a real robot and are most likely due to a combination of ``chance events'' that is very sensitive to simulation details. 
Overall, only 37.6\% of the simulated falls are realistically avoidable. As a consequence, we define a fall as ``avoidable'' if and only if the corresponding situation of wall configuration and robot posture leads to at least 9 (one position and its 8 neighbors on a grid) contiguous successful contact positions.
We focus on these ``avoidable situations'' as it is straightforward to generate an arbitrarily high number of non-avoidable situations and thus make the avoided fall rates arbitrarily low.

We randomly split this dataset into training (37.5\%), validation (12.5\%), and evaluation (50\%). To ensure that our results do not depend on a single, lucky run of the learning algorithms, we replicated the learning algorithm 20 times using 20 different splits of our dataset and different seeds.

%%%%%%%%%%%%%%%%%%%%%%%%%%%%%%%%%%%%%%%%%%%%%%%%%%%%%%%%%%%%
%%   Result
%%%%%%%%%%%%%%%%%%%%%%%%%%%%%%%%%%%%%%%%%%%%%%%%%%%%%%%%%%%%
\subsection{Results}

\begin{figure}[t]
    \includegraphics[width=\linewidth]{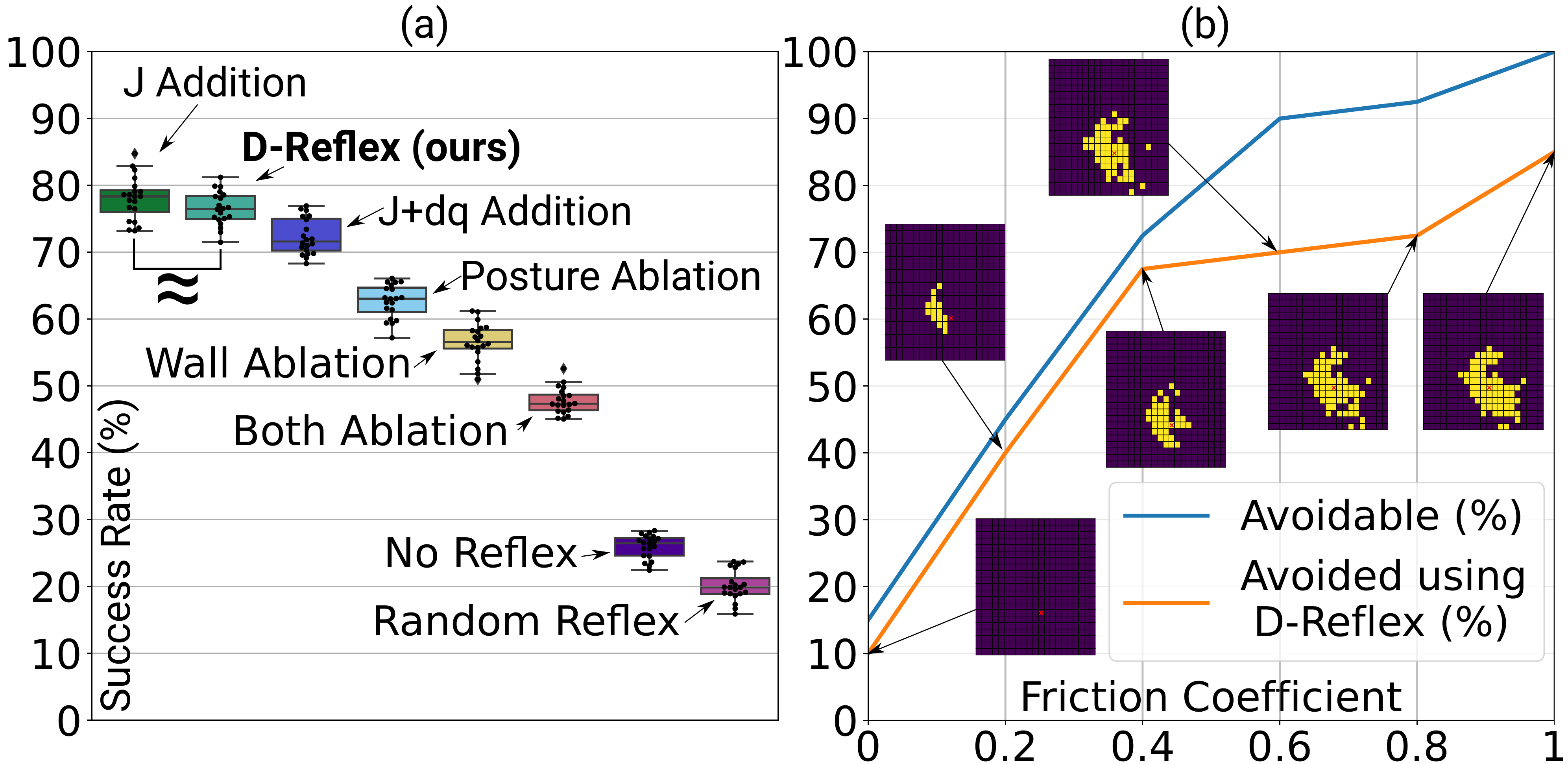}
    \caption{(a) Success rate considering only avoidable situations. The box plots show the median and quartiles, the bullets being the values of the evaluations. For each variant, the learning algorithm was run 20 times on different splits of the dataset. All methods are significantly different from one another using a t-test with Bonferroni correction (p-value $\le0.001$) except D-Reflex and the J Addition. (b) Percentage of avoidable situations and situations avoided using D-Reflex, with a model trained for the friction $1$, depending on the wall friction coefficient, plus examples of contact maps. }
    \label{fig:final_comparison}
   \vspace{-5mm}
   %\vspace{-1.1cm}
\end{figure}

Fig.~\ref{fig:main_figure} shows many examples of successful balance recoveries, as well as examples of typical failures; Fig.~\ref{fig:final_comparison} shows the comparison of the success rate for the avoidable conditions of our method against the different ablations, additions, and baselines. All methods are significantly different from each other, except the D-Reflex and the J Addition. The video (\url{https://youtu.be/hbuWr-ZNAtg}) presents more examples.

%%%%%%%%%%%%%%%%%%%%%%%%%%%%%%%%%%%%%%%%%%%%%%
%%%   Our method
%%%%%%%%%%%%%%%%%%%%%%%%%%%%%%%%%%%%%%%%%%%%%%
\subsubsection{Our method}
D-Reflex allows the TALOS robot to successfully recover from the fall for $76.4\% \pm 2.5\%$ of the avoidable conditions, Fig. \ref{fig:final_comparison}.
Fig.~\ref{fig:main_figure} showcases some examples of successful recoveries as well as some typical failures. Failure cases mostly happen when the neural network overestimates the size and the smoothness of the possible contact zone (Fig.~\ref{fig:main_figure}, fifth row); more tuning of the learning process or using another regression technique (e.g., random forests) could lead to successful contact zones that are less smooth and more accurate, with the same training data. Overall, our approach significantly (t-test with p-value $\le 0.001$) increases the likelihood of avoiding the fall and thus prevents the robot from being more damaged (which could be very costly) and allows it to finish its mission despite the damage (especially if the robot has access to a locomotion strategy that can deal with a damaged leg).

%%%%%%%%%%%%%%%%%%%%%%%%%%%%%%%%%%%%%%%%%%%%%%
%%%   Baselines
%%%%%%%%%%%%%%%%%%%%%%%%%%%%%%%%%%%%%%%%%%%%%%
\subsubsection{Baselines}
Doing nothing (without any reflex) results in a low success rate ($26.4\% \pm 1.7\%$ of the avoidable falls). It is not zero because some situations do not result in a fall (e.g. a locked ankle). Using a random contact position results in a lower success rate ($19.8\% \pm 2.2\%$ of the avoidable falls) because moving the arm randomly creates unnecessary motions that decrease the stability of the robot. These two cases show the need for a reflex.

%%%%%%%%%%%%%%%%%%%%%%%%%%%%%%%%%%%%%%%%%%%%%%
%%%   Ablations
%%%%%%%%%%%%%%%%%%%%%%%%%%%%%%%%%%%%%%%%%%%%%%
\subsubsection{Ablations}

When both the wall configuration and the robot posture are ignored to choose the contact point (``Both Ablation''), the robot avoids $47.3\% \pm 1.9\%$ (versus $76.4\%$ for D-Reflex) of the avoidable falls. This corresponds to choosing the best point of the wall on average on the training set, which is roughly in the middle of the wall. This demonstrates that the neural network does more than choose the same ``average position''. 
Both the wall configuration and the robot posture are necessary information for the classifier; of the two, the wall configuration is the most important (Fig. \ref{fig:final_comparison}).

%%%%%%%%%%%%%%%%%%%%%%%%%%%%%%%%%%%%%%%%%%%%%%
%%%   Additions
%%%%%%%%%%%%%%%%%%%%%%%%%%%%%%%%%%%%%%%%%%%%%%
\subsubsection{Additions}

Adding the faulty joints information (``J Addition'') does not significantly improve the performance and requires a stronger assumption about our knowledge of the damage condition. Adding the joints velocities (``$J+\mathrm{d}q$ Addition'') significantly reduces the performance; as the robot does not perform dynamic motions, adding uninformative inputs only decreases the neural network performance (which could be corrected by re-optimizing the hyper-parameters).

%%%%%%%%%%%%%%%%%%%%%%%%%%%%%%%%%%%%%%%%%%%%%%
%%%   Friction
%%%%%%%%%%%%%%%%%%%%%%%%%%%%%%%%%%%%%%%%%%%%%%

\subsubsection{Friction}

DART simulates the contact using the rigid contact dynamics, taking into account friction, bouncing, and restitution, which can be parameterized to simulate different materials. In our cases, the friction has a default coefficient of $1$ and a restitution coefficient of $0$. Decreasing the friction coefficient removes successful contact positions, which decreases the percentage of avoidable falls. A D-Reflex trained with a friction of $1$ generalizes to lower frictions down to 0.4 (Fig. \ref{fig:final_comparison}b) without degrading too much its performance, from $85\%$ to $67.5\%$. The decrease in performance comes from the large reduction of successful contact points when the friction is lower than 0.4 (Fig. \ref{fig:final_comparison}b). During our experiments on the real robot, we did not notice a significant discrepancy between simulation and reality regarding  friction.

%%%%%%%%%%%%%%%%%%%%%%%%%%%%%%%%%%%%%%%%%%%%%%%%%%%%%%%%%%%%
%%   Use of the perfect model
%%%%%%%%%%%%%%%%%%%%%%%%%%%%%%%%%%%%%%%%%%%%%%%%%%%%%%%%%%%%
\subsection{Use of the updated model in the whole-body controller}

In these experiments, the whole-body controller uses the model of the intact robot to compute the joint angles of the damaged one: we assumed that the model of the damaged robot was not known during the fall avoidance phase and therefore we could not use an updated model. Nevertheless, as a baseline, we checked what happens when we use the model of the damaged robot when generating the dataset; we replicated the learning 25 times, but we only considered the amputation of the knee.

The results are statistically equivalent to those obtained when the controller uses the model of the intact robot (median fall avoidance rate: $80.7\% \pm 2.5\%$ %$80.7\% [77.8\%, 82.7\%]$}
for the intact model versus $82.3\% \pm 2.4\%$ %$82.3\% [78.5\%, 83.8\%]$
with the updated model). This lack of difference is due to the fact that our method looks at the result of a simulation to decide on the success of a particular location on the wall, and not at how well the robot performs an expected motion. Put differently, a controller with an updated model is likely to put the robot's hand closer to the target location, whereas it often misses this target when using the model of the intact robot; but this does not matter because what is stored is the success (fall avoidance) for a target, regardless of where this target actually leads on the wall. This means that the whole-body controller is used to generate a trajectory and the simulator tells the algorithm whether this trajectory is successful or not: generating this trajectory differently does not fundamentally change the result. Similarly, a planning or model-predictive control algorithm \cite{talos_torque_control,mini_cheetah_falling} could be used instead of our whole-body controller to generate these trajectories: we do not expect any significant change in the result because an algorithm will still be needed to choose \emph{where} to put the hand, which is the problem solved by D-reflex.

%%%%%%%%%%%%%%%%%%%%%%%%%%%%%%%%%%%%%%%%%%%%%%%%%%%%%%%%%%%%
%%   Experiments on the real robot
%%%%%%%%%%%%%%%%%%%%%%%%%%%%%%%%%%%%%%%%%%%%%%%%%%%%%%%%%%%%
\subsection{Experiments on the real robot}

In these experiments, we placed the robot in a known location with respect to a wall with a mattress (to be able to repeat falls with and without D-Reflex) and set it in a known posture. We cut the power of the knee's actuator. The loss of power is detected and triggers the D-Reflex. Overall, in three out of the four situations that we tested, the robot successfully avoided the fall, whereas it fell when no reflex was triggered (see the video \url{https://youtu.be/hbuWr-ZNAtg}). These experiments demonstrate that the robot is fast enough to execute the learned reflex and that the learned solutions are robust enough to cross the reality gap most of the time. The onboard CPU computes the contact location using the classifier on average in 16.7ms [min:14.0ms, max:21.9ms], but a more recent CPU and/or a GPU can achieve better performance.

%%%%%%%%%%%%%%%%%%%%%%%%%%%%%%%%%%%%%%%%%%%%%%%%%%%%%%%%%%%%%%%%%%%%%%%%%%%%%%%%
%   Conclusion
%%%%%%%%%%%%%%%%%%%%%%%%%%%%%%%%%%%%%%%%%%%%%%%%%%%%%%%%%%%%%%%%%%%%%%%%%%%%%%%%
\section{Conclusion}

Given the robot posture and the configuration (distance and orientation) of a wall within arm's reach, our method, D-Reflex, uses a trained neural-network classifier to estimate a contact position that allows the robot to avoid the fall in more than 75\% of the avoidable cases. Our method is robust 
to different combinations of damage conditions (locked actuators, passive actuators, and amputation) and does not require the knowledge of the correct damaged model of the robot. It relies on a whole-body controller \cite{dalin_whole-body_2021} and sensory information that is easy to acquire.

Once stabilized, the robot will need to update its model and its position so that it can be controlled again. Numerous model identification methods can be investigated \cite{model_identification}. The robot can then continue its mission despite being damaged.

The main assumption of this work is that there exists a wall next to the robot. In future work, we will extend our method to select target contact positions for the hands in more complex environments (other walls, furniture, ...). 
In that case, we would not be able to cover the possible contact positions with a simple 2D grid: the main challenge will be to build a dataset that includes successful contact positions.

D-Reflex is designed to avoid falls in complex, high-risk/high-gain missions with humanoid robots. In such situations, one would expect the operator to be careful, avoid highly-dynamic motion, and keep the robot in static balance. Future work will extend the training set with more dynamic motions, like dynamic walking and high-speed movements, and with single-support motions.

\addtolength{\textheight}{-0.cm}   % This command serves to balance the column lengths
                                  % the text-height of the last page by a suitable amount.
                                  % sure that you do not shorten the text-height too much.

%%%%%%%%%%%%%%%%%%%%%%%%%%%%%%%%%%%%%%%%%%%%%%%%%%%%%%%%%%%%%%%%%%%%%%%%%%%%%%%%

% %%%%%%%%%%%%%%%%%%%%%%%%%%%%%%%%%%%%%%%%%%%%%%%%%%%%%%%%%%%%%%%%%%%%%%%%%%%%%%%%
% %   APPENDIX


% %%%%%%%%%%%%%%%%%%%%%%%%%%%%%%%%%%%%%%%%%%%%%%%%%%%%%%%%%%%%%%%%%%%%%%%%%%%%%%%%
% %   ACKNOWLEDGMENT
% %%%%%%%%%%%%%%%%%%%%%%%%%%%%%%%%%%%%%%%%%%%%%%%%%%%%%%%%%%%%%%%%%%%%%%%%%%%%%%%%
% \section*{ACKNOWLEDGMENT}

%%%%%%%%%%%%%%%%%%%%%%%%%%%%%%%%%%%%%%%%%%%%%%%%%%%%%%%%%%%%%%%%%%%%%%%%%%%%%%%%

\bibliographystyle{IEEEtran}
\bibliography{IEEEabrv, bibliography}

\begin{thebibliography}{10}
\providecommand{\url}[1]{#1}
\csname url@rmstyle\endcsname
\providecommand{\newblock}{\relax}
\providecommand{\bibinfo}[2]{#2}
\providecommand\BIBentrySTDinterwordspacing{\spaceskip=0pt\relax}
\providecommand\BIBentryALTinterwordstretchfactor{4}
\providecommand\BIBentryALTinterwordspacing{\spaceskip=\fontdimen2\font plus
\BIBentryALTinterwordstretchfactor\fontdimen3\font minus
  \fontdimen4\font\relax}
\providecommand\BIBforeignlanguage[2]{{%
\expandafter\ifx\csname l@#1\endcsname\relax
\typeout{** WARNING: IEEEtran.bst: No hyphenation pattern has been}%
\typeout{** loaded for the language `#1'. Using the pattern for}%
\typeout{** the default language instead.}%
\else
\language=\csname l@#1\endcsname
\fi
#2}}

\bibitem{humanoids_ref}
A.~Goswami and P.~Vadakkepat, \emph{Humanoid Robotics: A Reference},
  1st~ed.\hskip 1em plus 0.5em minus 0.4em\relax Springer Publishing Company,
  Incorporated, 2018.

\bibitem{DRC_paper}
C.~Atkeson \emph{et~al.}, ``What happened at the {DARPA} robotics challenge
  finals,'' in \emph{The DARPA Robotics Challenge Finals: Humanoid Robots To
  The Rescue}, 2018, pp. 667--684.

\bibitem{model_identification}
J.~Hollerbach, W.~Khalil, and M.~Gautier, \emph{Model Identification}.\hskip
  1em plus 0.5em minus 0.4em\relax Berlin, Heidelberg: Springer Berlin
  Heidelberg, 2008, pp. 321--344.

\bibitem{talos_actuator_identification}
N.~Ramuzat \emph{et~al.}, ``Actuator model, identification and differential
  dynamic programming for a talos humanoid robot,'' in \emph{European Control
  Conference (ECC)}, 2020, pp. 724--730.

\bibitem{qp_repulsors}
J.~Spitz \emph{et~al.}, ``Trial-and-error learning of repulsors for humanoid
  {QP}-based whole-body control,'' in \emph{IEEE Humanoids}, 2017.

\bibitem{mine_meta_learning}
T.~Anne, J.~Wilkinson, and Z.~Li, ``Meta-learning for fast adaptive locomotion
  with uncertainties in environments and robot dynamics,'' in \emph{{IEEE/RSJ}
  IROS}, 2021, pp. 4568--4575.

\bibitem{rituraj_meta_learning}
R.~Kaushik, T.~Anne, and J.~Mouret, ``Fast online adaptation in robotics
  through meta-learning embeddings of simulated priors,'' in \emph{{IEEE/RSJ}
  IROS}.\hskip 1em plus 0.5em minus 0.4em\relax {IEEE}, 2020, pp. 5269--5276.

\bibitem{nature_paper}
A.~Cully, J.~Clune, D.~Tarapore, and J.~Mouret, ``Robots that can adapt like
  animals,'' \emph{Nature}, vol. 521, no. 7553, pp. 503--507, 2015.

\bibitem{nagabandi_meta_learning}
A.~Nagabandi, I.~Clavera, S.~Liu, R.~S. Fearing, P.~Abbeel, S.~Levine, and
  C.~Finn, ``Learning to adapt in dynamic, real-world environments through
  meta-reinforcement learning,'' in \emph{ICLR}, 2019.

\bibitem{residual_paper2}
K.~Chatzilygeroudis and J.-B. Mouret, ``Using parameterized black-box priors to
  scale up model-based policy search for robotics,'' in \emph{IEEE ICRA}, 2018.

\bibitem{bouyarmane2017weight}
K.~Bouyarmane and A.~Kheddar, ``On weight-prioritized multitask control of
  humanoid robots,'' \emph{IEEE Transactions on Automatic Control}, vol.~63,
  no.~6, pp. 1632--1647, 2017.

\bibitem{dalin_whole-body_2021}
E.~Dalin \emph{et~al.}, ``{Whole-body teleoperation of the Talos humanoid
  robot: preliminary results},'' in \emph{{ICRA 2021 Workshop on
  Teleoperation}}.

\bibitem{talos_ref}
O.~Stasse \emph{et~al.}, ``Talos: A new humanoid research platform targeted for
  industrial applications,'' in \emph{{IEEE} Humanoids}, 2017.

\bibitem{kheddar2019humanoid}
A.~Kheddar \emph{et~al.}, ``Humanoid robots in aircraft manufacturing: The
  {Airbus} use cases,'' \emph{IEEE RA Magazine}, vol.~26, no.~4, pp. 30--45,
  2019.

\bibitem{PolveriniHLT21}
M.~P. Polverini \emph{et~al.}, ``Agile actions with a centaur-type humanoid:
  {A} decoupled approach,'' in \emph{{IEEE} ICRA}, 2021.

\bibitem{Crocoddyl}
C.~Mastalli \emph{et~al.}, ``Crocoddyl: An efficient and versatile framework
  for multi-contact optimal control,'' in \emph{IEEE ICRA}, 2020.

\bibitem{cui_human_2021}
D.~Cui \emph{et~al.}, ``Human inspired fall arrest strategy for humanoid robots
  based on stiffness ellipsoid optimisation,'' \emph{Bioinspiration \&
  biomimetics}, vol.~16, 2021.

\bibitem{kheddar1}
V.~Samy and A.~Kheddar, ``Falls control using posture reshaping and active
  compliance,'' in \emph{{IEEE-RAS} Humanoids}, 2015.

\bibitem{fall_minimization_back}
Q.~Li \emph{et~al.}, ``A minimized falling damage method for humanoid robots,''
  \emph{IJARS}, vol.~14, no.~5, 2017.

\bibitem{Liu21}
D.~Liu, Y.~Lin, and V.~Kapila, ``A rollover strategy for wrist damage reduction
  in a forward falling humanoid,'' in \emph{IEEE International Conference on
  Mechatronics and Automation (ICMA)}, 2021.

\bibitem{kheddar2}
V.~Samy, K.~Bouyarmane, and A.~Kheddar, ``{QP}-based adaptive-gains compliance
  control in humanoid falls,'' in \emph{IEEE ICRA}, 2017.

\bibitem{kheddar3}
V.~Samy, S.~Caron, K.~Bouyarmane, and A.~Kheddar, ``Post-impact adaptive
  compliance for humanoid falls using predictive control of a reduced model,''
  in \emph{{IEEE-RAS} Humanoids}, 2017.

\bibitem{mini_cheetah_falling}
V.~Kurtz, H.~Li, P.~M. Wensing, and H.~Lin, ``Mini cheetah, the falling cat:
  {A} case study in machine learning and trajectory optimization for robot
  acrobatics,'' \emph{CoRR}, vol. abs/2109.04424, 2021.

\bibitem{RANSAC}
M.~Fischler and R.~Bolles, ``Random sample consensus: A paradigm for model
  fitting with applications to image analysis and automated cartography,''
  \emph{Communications of the ACM}, vol.~24, no.~6, 1981.

\bibitem{SPD}
J.~Tan, K.~Liu, and G.~Turk, ``Stable proportional-derivative controllers,''
  \emph{IEEE Computer Graphics and Applications}, vol.~31, no.~4, 2011.

\bibitem{lee2018dart}
J.~Lee \emph{et~al.}, ``Dart: Dynamic animation and robotics toolkit,''
  \emph{Journal of Open Source Software}, vol.~3, no.~22, p. 500, 2018.

\bibitem{talos_torque_control}
E.~L. Dantec \emph{et~al.}, ``Whole body model predictive control with a memory
  of motion: Experiments on a torque-controlled {Talos},'' in \emph{IEEE ICRA},
  2021.

\end{thebibliography}

\end{document}